\pdfoutput=1

\documentclass[11pt]{article}

\usepackage[preprint]{coling}

\usepackage{times}
\usepackage{latexsym}

\usepackage[T1]{fontenc}

\usepackage[utf8]{inputenc}

\usepackage{microtype}

\usepackage{inconsolata}

\usepackage{graphicx}

\title{Show Less, Instruct More:\\Enriching Prompts with Definitions and Guidelines for Zero-Shot NER}

\author{
    \textbf{Andrew Zamai \textsuperscript{1} }
    \textbf{Andrea Zugarini \textsuperscript{2} }
    \textbf{Leonardo Rigutini \textsuperscript{2}}
    \AND
    \textbf{Marco Ernandes \textsuperscript{2} }
    \textbf{Marco Maggini \textsuperscript{1}}
\\
\\
    \textsuperscript{1} University of Siena, Italy \hspace{0.5cm}
    \textsuperscript{2} expert.ai, Siena, Italy
\\
\small{
    \{andrew.zamai, marco.maggini\}@unisi.it
    }
\\
\small{
    \{azugarini, lrigutini, mernandes\}@expert.ai
    }
}

\newcommand{\andrew}[1]{\color{black} #1}

\usepackage{amssymb}
\usepackage{booktabs}
\usepackage{soul} 
\usepackage{array}

\usepackage{color, colortbl}
\definecolor{LightCyan}{rgb}{0.88,1,1}
\definecolor{LightGreen}{rgb}{0.88,1,0.88}
\definecolor{LightRed}{rgb}{1,0.88,0.88}
\definecolor{LightGray}{rgb}{0.92,0.92,0.92}

\begin{document}

\maketitle

\begin{abstract}
Recently, several specialized instruction-tuned Large Language Models (LLMs) for Named Entity Recognition (NER) have emerged. Compared to traditional NER approaches, these models have demonstrated strong generalization capabilities. Existing LLMs primarily focus on addressing zero-shot NER on Out-of-Domain inputs, while fine-tuning on an extensive number of entity classes that often highly or completely overlap with test sets. In this work instead, we propose SLIMER, an approach designed to tackle never-seen-before entity tags by instructing the model on fewer examples, and by leveraging a prompt enriched with \textit{definition} and \textit{guidelines}.
Experiments demonstrate that definition and guidelines yield better performance, faster and more robust learning, particularly when labelling unseen named entities. Furthermore, SLIMER performs comparably to state-of-the-art approaches in out-of-domain zero-shot NER, while being trained in a more fair, though certainly more challenging, setting.
\end{abstract}

\begin{figure}[t]
\centering
  \includegraphics[width=0.95\columnwidth]{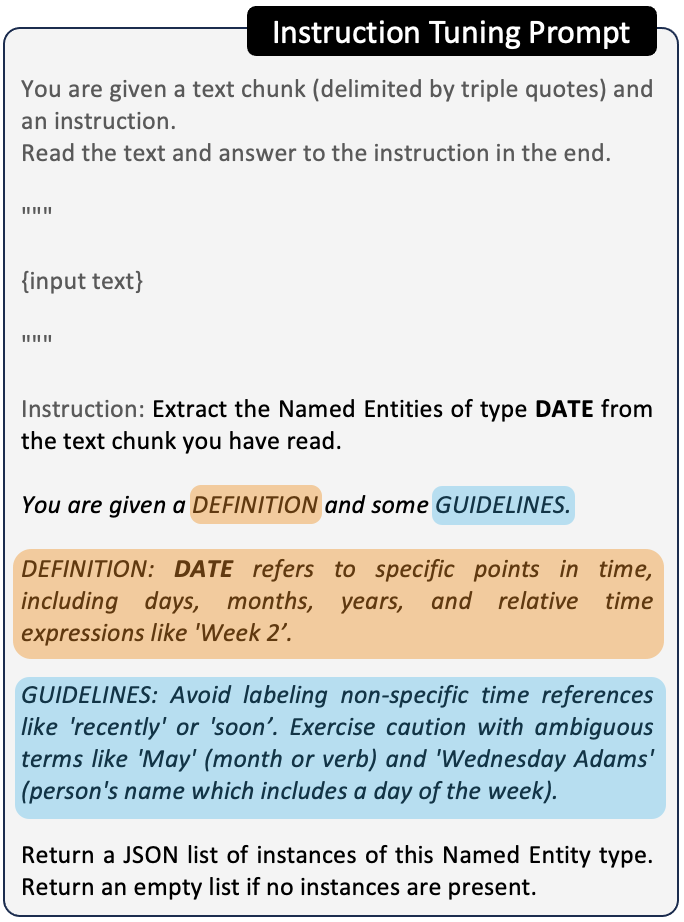}
  \caption{SLIMER's prompt. Dedicated entity definition and guidelines steer the model generation.
  }
  \label{fig:our_instruction_prompt}
\end{figure}

\section{Introduction}
Named Entity Recognition (NER) is a crucial problem in Natural Language Processing (NLP), usually being a key component in Information Extraction pipelines. Traditional methods frame NER into a sequence labeling task \cite{li2020survey}, where models are specialized on a narrow domain and a pre-defined label set, thus lacking generalization capabilities outside the downstream task at hand.
On the contrary, Large Language Models (LLMs) have demonstrated strong zero-shot capabilities.
Through cleverly designed prompts, models like GPT-3 can tackle NER via In-Context Learning \cite{radford2019language, brown2020language}. However, smaller (encoder-only) LMs, trained on the specific NER task, may still outperform LLMs~\citep{wang2023gpt, ye2023comprehensive, zhou2023universalner}. To this end, Instruction-Tuning of LLMs has emerged as an effective method to improve their performance \cite{wei2022finetuned, chung2022scaling, wang-etal-2022-super}. In the literature, several works  have explored Instruction-Tuning for NER, including InstructUIE \cite{wang2023instructuie}, UniNER \cite{zhou2023universalner}, GoLLIE \cite{sainz2024gollie} and GNER \cite{ding2024rethinking}.

\begin{table*}[ht]
  \centering
  \small
  \begin{tabular}{r|cccccc}
    \toprule
     \textbf{} & \textbf{InstructUIE} & \textbf{UniNER} & \textbf{GoLLIE} & \textbf{GLiNER} & \textbf{GNER} & \textbf{SLIMER (our)}\\
    \midrule
    Architecture & enc-dec & decoder & decoder & encoder & (enc-)dec & decoder\\
    
    NER-only & $\times$ & $\checkmark$ & $\times$ & $\checkmark$ & $\checkmark$ & $\checkmark$\\

    nested-NER & $\checkmark$ & $\checkmark$ & $\checkmark$ & $\times$ & $\times$ & $\checkmark$\\
    
    Instruction-tuned & $\checkmark$ & $\checkmark$ & $\checkmark$ & $\times$ & $\checkmark$ & $\checkmark$\\

    Instruction template & list tuples & conversation & Py-classes & $\times$ & gen-BIO & guided inst \\
    
    NE Guidelines & $\times$ & $\times$ & $\checkmark$ & $\times$ & $\times$ & $\checkmark$\\

    
    

    Inference cost & $|\mathcal{X}|$ & $|\mathcal{X}| \times |\mathcal{Y}|$& $|\mathcal{X}|$ & $|\mathcal{X}|$ & $|\mathcal{X}|$ & $|\mathcal{X}| \times |\mathcal{Y}|$\\

    Works document level & $\checkmark$ & $\checkmark$ & $\checkmark$ & $\checkmark$ & $\times$ & $\checkmark$\\

    Trained on synthetic data & $\times$ & $\checkmark$ & $\times$ & $\checkmark$ & $\checkmark$ & $\checkmark$\\

    \# distinct NEs & $\leq 119$ & $13020$ & $\leq 40$ & $13020$ & $13020$ & $391$\\

    Human effort for guidelines & $\times$ & $\times$ & high & $\times$ & $\times$ & gpt-prompt\\

    \midrule
    Out-Of-Domain evaluation & \checkmark & \checkmark & \checkmark & \checkmark & \checkmark & \checkmark \\
    Unseen NEs evaluation & $\times$ & $\times$ & \checkmark & $\times$ & $\times$ & \checkmark \\
    
    \bottomrule
  \end{tabular}
  \caption{
    Overview of existing works in the literature, highlighting the differences on some identified key comparative features. $|\mathcal{X}|$ denotes the number of text inputs, $|\mathcal{Y}|$ the number of NEs. The symbol $\leq$ is used to indicate the upper bound on the number of distinct NEs in training when there is no overlap of label sets between the merged datasets.
  }\label{tab:prev_works_comparison}
\end{table*}

Such dedicated LLMs have the ability to perform zero-shot NER on heterogeneous input domains and a multitude of possibly never-seen-before NEs.
Existing works mainly focus on zero-shot NER on Out-Of-Domain (OOD) inputs, while fine-tuning on an extensive number of entity classes that often highly or completely overlap between the training and test sets (see Table~\ref{tab:NEs_overlap}). Consequently, the problem of tagging unseen named entities has been little investigated, with GoLLIE as the only exception.

In this work instead, we tackle both scenarios by carefully selecting a training set with minimal degree of class-overlap with test data. To facilitate effective zero-shot capabilities on such novel NEs, we guide the model with dedicated definition and guidelines for the category to be annotated.
By using fewer training samples from a reduced number of distinct named entity tags, combined with prompts enriched of \textit{definition} and \textit{guidelines}, we name our approach \textbf{SLIMER}: \textbf{S}how \textbf{L}ess, \textbf{I}nstruct \textbf{M}ore - \textbf{E}ntity \textbf{R}ecognition\footnote{\url{https://huggingface.co/expertai/SLIMER}}.

Experiments were conducted on two standard NER benchmarks for zero-shot OOD, MIT~\citep{MITliu2013asgard} and CrossNER~\citep{liu2021crossner}. Additionally, we assessed performance on never-seen-before NEs on BUSTER \cite{zugarini2023buster}, a document-level NER dataset with entity tags novel to all the evaluated models. 
We explored how increasing both the number of training samples and the number of unique Named Entity (NE) categories affect the generalization capabilities of LLMs, with or without the support of definitions and guidelines.
Comparison of SLIMER with its baseline, i.e. the model devoid of definition and guidelines, reveals SLIMER's deeper understanding, faster and more stable learning, and better zero-shot performance. Despite being trained on a fraction of the data, with little overlap between train and test named entity tags, SLIMER performs comparably against state-of-the-art instruction-tuned approaches, revealing stronger generalization capabilities when dealing with unseen named entities.

\begin{table*}[!ht]
  \centering
  \small
  \resizebox{\textwidth}{!}{
  \begin{tabular}{l|cc|ccccc|c}
    \toprule
    \textbf{Model(s)} & \multicolumn{2}{c|}{\textbf{MIT}} & \multicolumn{5}{c|}{\textbf{CrossNER}} & \textbf{TOT}\\
    
    & Movie & Restaurant & AI & Literature & Music & Politics & Science &  \\
    \midrule
    
    UniNER, GLiNER, GNER & $10/12$ & $8/8$ & $13/13$ & $10/11$ & $12/12$ & $8/8$ & $16/16$ & $52/55$\\
    & $83\%$ & $100\%$ & $100\%$ & $91\%$ & $100\%$ & $100\%$ & $100\%$ &  $95\%$\\

    \midrule

    GoLLIE & $1/12$ & $3/8$ & $5/13$ & $5/11$ & $5/12$ & $6/8$ & $7/16$ & $12/55$\\
    & $8\%$ & $38\%$ & $38\%$ & $45\%$ & $42\%$ & $75\%$ & $44\%$ & $22\%$\\

    \midrule
    
    SLIMER & $0/12$ & $1/8$ & $4/13$ & $4/11$ & $4/12$ & $4/8$ & $4/16$ & $4/55$\\
    & $0\%$ & $13\%$ & $31\%$ & $36\%$ & $33\%$ & $50\%$ & $25\%$ & $7\%$\\
    
    \bottomrule
  \end{tabular}
}
  \caption{Overlap (\%) between the NE types seen during training and evaluation by each model in MIT and CrossNER benchmarks. Since some classes are shared among different evaluation datasets, the final column accounts for the fraction relative to unique entity types.
    UniNER, GLINER, and GNER models exhibit near-complete overlap, meaning that in practice there are no unseen named entity classes in their test sets.
  }\label{tab:NEs_overlap}
\end{table*}

\section{Related Work}\label{sec:related_works}
Commonly employed machine-learning solutions frame NER into a ``sequence labeling task'', where the goal is to assign a BIO label to each element in a given sequence \cite{li2020survey}. Fine-tuning BERT-family models~\citep{devlin2018bert} for NER is a well established approach.
While these models excel in supervised contexts, they have the severe limitation of being constrained to a predefined set of labels and inputs from limited domains, making it difficult for them to generalize across different contexts and on unseen named entities.

\subsection{In-Context Learning}


LLMs have demonstrated impressive few-shot and zero-shot capabilities on various challenging tasks \cite{radford2019language, brown2020language}. However, endeavours to utilize LLMs for Information Extraction have been less promising \cite{keraghel2024survey}. Attempts to make use of LLMs (such as GPT) through clever prompt engineering have been conducted by \citet{wang2023gpt} in their paper GPT-NER. 
In \citet{ye2023comprehensive}, the authors compared several GPT models on various NLU tasks, including NER. The results highlighted a significant gap compared to supervised encoder-only approaches. Those results were further confirmed by \citet{zhou2023universalner} on other NER datasets.

\subsection{Fine-Tuning for Zero-Shot NER}
Parallel to In-Context Learning, other approaches explored Instruction-Tuning \cite{wei2022finetuned, chung2022scaling, wang-etal-2022-super}, focusing the training on a dedicated task, such as NER. 

InstructUIE \cite{wang2023instructuie} is a encoder-decoder T5-11B model fine-tuned on supervised IE datasets (among which NER) phrased as text-to-text problems, the model is instructed to return a list of tuples \textit{(text span, entity type)}, choosing from a given list of categories. 
UniNER \cite{zhou2023universalner} consists in a decoder-only LLaMA model fine-tuned on a ``conversation style template''. In inference, the model is prompted with the question ``What describes NE in the text?'' and a list of entities that belong to the requested NE is returned.

Several other works have followed, each based on a different instruction-tuning template.
Table~\ref{tab:prev_works_comparison} provides an overview over the most significant state-of-the-art instruction-tuned models for NER, highlighting their architecture, the designed prompt template, and other key comparative features. 
GNER \cite{ding2024rethinking} rethinks the importance of negative instances (i.e., ``O'' tags in BIO labeling) and replaces the established entity-centric schema with a BIO-like generation, replicating the same input text along with token-by-token BIO labels. Despite input length limitations and output parsing difficulties, their approach displays strong boundaries detection and reduced classification indecision.
GLiNER~\cite{zaratiana2023gliner} instead, relies on a much smaller, encoder-only and non-instruction-tuned architecture, which achieves remarkable performance in both supervised and zero-shot NER.

\paragraph{Definitions and guidelines for NER.}
There have been some attempts to instruct models with additional information about the entity type(s).
\citet{zhou2023universalner} experimented an UniNER variant enriched with brief entity definitions. However, such definitions were used in replacement of the entity name, rather than being provided as a supplementary aid. Generated contextually to a synthetic annotation process, they often varied for the same entity type, thus lacking of consistency, structure, and richness of content. As a result, \textit{UniNER-def} model under-performed compared to its original version.

The effectiveness of adding NER tags descriptions was demonstrated in other works~\cite{sainz2024gollie, label_verbalization_NER}. 
In particular, GoLLIE's authors \citet{sainz2024gollie}, further explored this direction by adopting a code-based representation where the NE labels are encoded as Python classes, and guidelines are formatted as doc-strings. Comparison with a baseline excluding them demonstrated their effectiveness in tagging unseen NEs.

In this work, alike GoLLIE, we steer the model with annotation guidelines. However, we replace their code-based representation with a more natural language prompt that includes a \textit{definition} and some \textit{guidelines} for the category to be extracted. Our guidelines are longer, richer and more descriptive, yet simpler in syntax, making them a more intuitive and accessible solution for non-technical users who do not need to deal with code.
Moreover, as also noted by GoLLIE's authors, code-based syntax is more effective on code-oriented backbones. In contrast, these models are less familiar with long natural language sentences. 
This hinders the adoption of GoLLIE's framework on a broad range of LLMs.



\section{SLIMER}
This section presents our approach, SLIMER. First, we provide motivations for reducing the number of training samples. Then, we describe the instruction-tuning prompt. Finally, we discuss how one can generate definitions and guidelines automatically by means of another LLM, such as ChatGPT.

\subsection{Show Less}
Existing models for zero-shot NER are trained on a large set of entity tags and examples. This training data can be generated synthetically ~\citep{zhou2023universalner}, by merging existing human-labelled datasets ~\citep{sainz2024gollie}, or even combing the two approaches ~\citep{zhou2023universalner}. While training on such extensive data certainly strengthens cross-domain zero-shot NER performance, it is unclear how it affects generalization capabilities on never-seen-before entity tags. Furthermore, as already observed in literature, instruction tuning helps aligning the model to the task and desired output format, but most of the gains in performance are achievable with little amounts of instructions~\citep{wang2022self, zhou2024lima, zugarini-etal-2024-clue-instruct}. 
Motivated by this, we train SLIMER on a fraction of the training data that is typically used to instruction-tune zero-shot models for NER.

\subsection{Instruct More}
As we reduce the data, in contrast, we enrich the model prompt with a \textit{definition} and some \textit{guidelines} about the entity to tag.
An example of instruction tuning prompt is illustrated in Figure~\ref{fig:our_instruction_prompt}. 

\paragraph{Definition and Guidelines.} The definition for a NE is designed to be a short sentence describing the tag at hand. This is followed by guidelines that provide annotation directives to align the model's labeling with the desired annotation scheme. Guidelines can be used to discourage the model from labelling particular edge cases or to provide examples of such NE. Thus, these components are intended to better instruct the model with specifics that well define what to extract and what not to extract. Moreover, such an information is crucial when dealing with unfamiliar entity tags, and it also allows to distinguish between polysemous categories.
From now on, we refer to Definition and Guidelines as D\&G.

\paragraph{Prompt structure.} According to \citet{zhou2023universalner}, we designed the prompt to extract the occurrences of one entity type per call. This has the drawback of requiring $|\mbox{NE}|$ inference calls on each input text, but allows the model to better focus on a single NE type at the time. Moreover, compared to GoLLIE, where all tags's guidelines are prepended to the input, each individual instruction will be simpler and shorter. 

\paragraph{Output.} We ask the model to generate its output in a parsable JSON format consisting of a list of NE instances identified in the given input. 
It is worth noticing that, in the LLM fine-tuning the Next To Prediction-loss penalizes the order of the returned tokens. Hence, during training we sort the target entities by their order of appearance within the input text. Moreover, since it's redundant to return the same instance text multiple times, we reduce the list of target instances to a set of unique text instances.

\subsection{Definition and Guidelines generation}
To fully exploit the potential benefits of D\&G, we must have high-quality information about an entity.

\begin{figure}[ht!]
    \centering
    \includegraphics[width=0.8\columnwidth]{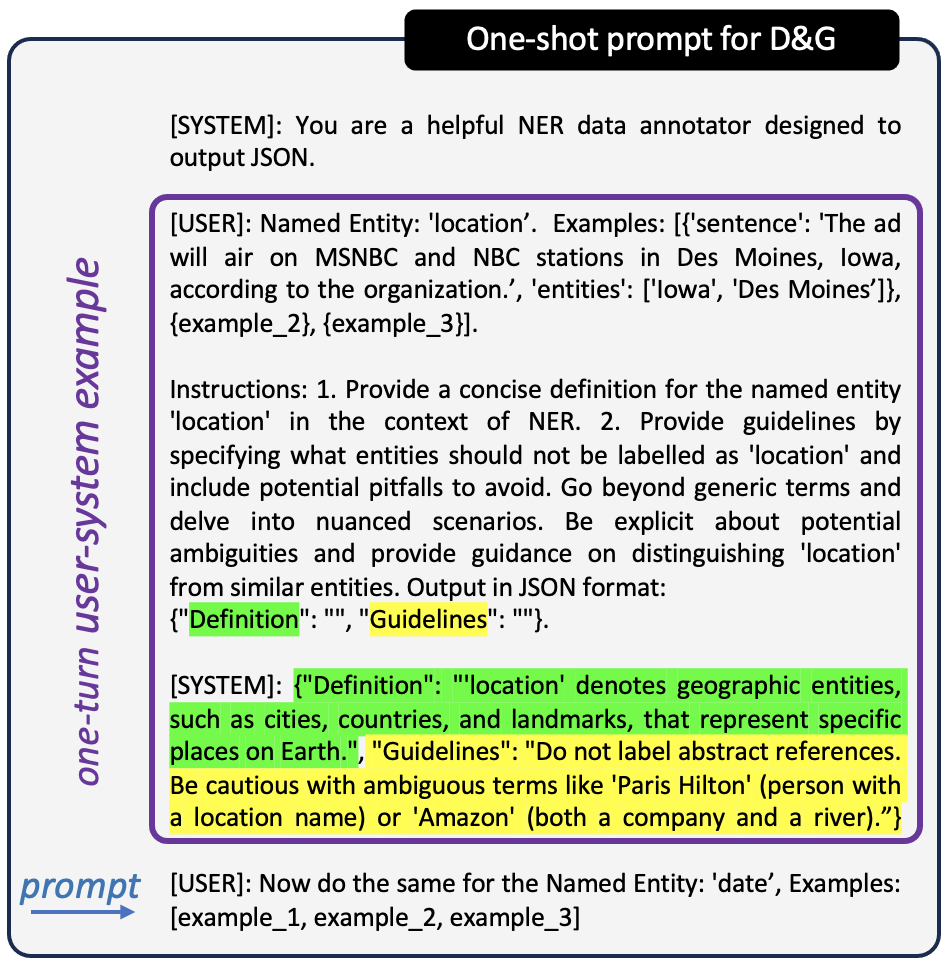}
    \caption{Prompt for generating the Definition and the Guidelines for a specific named entity.}
    \label{fig:guidelines_prompt}
\end{figure}

When the number of entity types is small, their production can be tackled manually, but as the number grows, it may require excessive human effort, as also pointed out in GoLLIE~\cite{sainz2024gollie}. 

To overcome this limitation, we exploited the OpenAI's Chat-GPT APIs to automatically generate definition and guidelines.
In particular, we designed the \textit{one-shot prompt template} reported in Figure~\ref{fig:guidelines_prompt}, which was used to query gpt-3.5-turbo-1106. 
An exemplary one-round user-system conversation is used to illustrate the desired output to the model. 
The three examples are randomly sampled for each NE from the dataset at hand.
Thanks to such a prompt, all the generated definitions and guidelines exhibit a similar structure, with short and clear defining sentences, and with guidelines highlighting edge cases where to be cautious\footnote{Some definition and guidelines examples in Appendix~\ref{app:appendix_some_examples}.}.

\section{Experiments}
In the experiments, we investigate the impact of definition and guidelines on zero-shot NER. We compare SLIMER against state-of-the-art models on OOD inputs and unseen entity types. 

\subsection{Datasets}\label{subsec:exps_datasets}
SLIMER models have all been fine-tuned on a subset of PileNER-type~\citep{zhou2023universalner}, and evaluated on three different benchmarks: MIT~\citep{MITliu2013asgard}, CrossNER~\citep{liu2021crossner} and BUSTER~\citep{zugarini2023buster}

\paragraph{PileNER-type.} PileNER-type~\citep{zhou2023universalner} is a synthetic dataset comprising a large set of approximately 50 thousands examples, encompassing over 13 thousands different named entity types. 
We kept only a subset of them, considering only those NEs with at least 100 instances. From the remaining 455 named entity classes, we manually revised and merged together classes of identical types spelled differently, e.g. \textit{organisation} and \textit{organization}, and discarded some ``catch-all'' gpt-hallucinated labels (e.g. \textit{unknown, other, miscellaneous, general, entity type}), thus further reducing to 423 different labels. 
Finally, to limit the overlap between training and test entity types, we excluded nearly all the categories present in the test datasets, with the exception of the standard NER tags:  \textit{person, location, organization, country}. Overall, we kept 391 distinct NEs. Percentages of overlap between train and test are reported in Table~\ref{tab:NEs_overlap}. 
Further details about the dataset composition, grouped by topic, are illustrated in the appendix (see Figure~\ref{fig:macro-topics}).

\paragraph{MIT and CrossNER.} MIT~\citep{MITliu2013asgard} and CrossNER~\citep{liu2021crossner} datasets have become de-facto the standard benchmark for zero-shot NER. We use them to compare SLIMER against existing state-of-the-art models on out-of-domain (OOD) inputs.

\paragraph{BUSTER.} We extend the evaluation beyond the MIT and CrossNER benchmarks by including BUSTER~\citep{zugarini2023buster}. 
The dataset is a document-level NER benchmark in financial domain. Both domain and named entity tags differ from what observed by all the models during instruction-tuning, i.e. have 0\% overlap. Its significant differences from standard NER datasets make BUSTER a perfect benchmark for evaluating zero-shot performance on never-seen-before NEs.


\begin{table*}[htb]
\centering
\footnotesize
\resizebox{\textwidth}{!}{
\begin{tabular}{l|cc|ccccc|c|c}
    \toprule
    
    \textbf{Model} &
    \multicolumn{2}{c|}{\textbf{MIT}} & \multicolumn{5}{c|}{\textbf{CrossNER}} & & \textbf{seen NEs}\\

     & Movie & Restaurant & AI & Literature & Music & Politics & Science & AVG & \\
    
    \midrule
    
    GPT-3.5-turbo & 5.3 & 32.8 & 52.4 & 39.8 & 66.6 & 68.5 & 67.0 & 47.5 & - \\

    \midrule
    
    InstructUIE & 63.0 & 21.0 & 49.0 & 47.2 & 53.2 & 48.2 & 49.3 & 47.3 & 20\% \\

    UniNER-type & 42.4 & 31.7 & 53.5 & 59.4 & 65.0 & 60.8 & 61.1 & 53.4 & \textcolor{red}{95\%} \\
    UniNER-def & 27.1 & 27.9 & 44.5 & 49.2 & 55.8 & 57.5 & 52.9 & 45.0 & \textcolor{red}{$\geq$ 95\%}\\
    UniNER-type+sup. & 61.2 & 35.2 & 62.9 & 64.9 & 70.6 & 66.9 & 70.8 & 61.8 & \textcolor{red}{$\geq$ 95\%}\\
    
    GoLLIE & 63.0 & 43.4 & 59.1 & 62.7 & 67.8 & 57.2 & 55.5 & 58.4 & 22\% \\ 

    GLiNER-L & 57.2 & 42.9 & 57.2 & 64.4 & 69.6 & 72.6 & 62.6 & 60.9 & \textcolor{red}{95\%} \\ 

    GNER-T5 & 62.5 & 51.0 & 68.2 & 68.7 & 81.2 & 75.1 & 76.7 & 69.1 & \textcolor{red}{95\%} \\

    GNER-LLaMA & 68.6 & 47.5 & 63.1 & 68.2 & 75.7 & 69.4 & 69.9 & 66.1 & \textcolor{red}{95\%} \\

    \midrule
    
    SLIMER w/o D\&G & $46.4 \pm {1.8}$ & $36.3\pm{2.1}$ & $49.6\pm{3.2}$ & $58.4\pm{1.7}$ & $56.8\pm{2.1}$ & $57.9\pm{2.1}$ & $53.8\pm{1.7}$ & $51.3 \pm {2.0}$ & \textcolor{blue}{7\%}\\
    
    \rowcolor{LightGray}
    SLIMER & $50.9\pm0.9$ & $38.2\pm0.3$ & $50.1\pm2.4$ & $58.7\pm0.2$ & $60.0\pm0.5$ & $63.9\pm1.0$ & $56.3\pm0.6$ & $54.0\pm0.5$ & \textcolor{blue}{7\%}\\

    \bottomrule
    \end{tabular}
}
    \caption{
    \small{
    Performance comparison between SLIMER and state-of-the-art models on the MIT and CrossNER benchmarks.
    Most models operate in an out-of-domain setting, due to their almost complete overlap between train and test tag sets. SLIMER instead tackles a 
    more challenging scenario, with a significantly reduced train-test classes sharing. 
    With the exception of UniNER-def, all the competitors' results are taken from their respective papers as listed in Section~\ref{sec:compared_models}.
    }
    }
    \label{tab:MIT_CrossNER_comparison}

\end{table*}

\begin{table*}[htb]
\centering
\footnotesize
\resizebox{\textwidth}{!}{
\begin{tabular}{llc|cc|ccccc|c||c}
    \toprule
    
    
    \textbf{Model} & \textbf{Backbone} & \textbf{Guidelines} &
    \multicolumn{2}{c|}{\textbf{MIT}} & \multicolumn{5}{c|}{\textbf{CrossNER}} & & \textbf{BUSTER}\\

    \textbf{} & \textbf{} & & Movie & Restaurant & AI & Literature & Music & Politics & Science & AVG & \textbf{} \\
    \midrule
    
    GNER & LLaMA2-chat-7B & $\times$ & 37.1 & 34.3 & 47.1 & 50.4 & 60.4 & 52.7 & 55.3 & 48.2 & 14.4 \\

    

    GoLLIE & LLaMA2-chat-7B & $\checkmark$ & 47.9 & \textbf{39.2} & \textbf{54.2} & 51.4 & \textbf{62.5} & 61.9 & 55.0 & 53.2 & 15.2 \\

    

    SLIMER & LLaMA2-chat-7B & $\checkmark$ & \textbf{50.9} & 38.2 & 50.1 & \textbf{58.7} & 60.0 & \textbf{63.9} & \textbf{56.3} & \textbf{54.0} & \textbf{45.3}\\

    \bottomrule
    \end{tabular}
}
    \caption{Comparing SLIMER against state-of-the-art models when provided same LLM backbone and fine-tuning on the same sub-set of unique Named Entity types, such that the overlap between train-test sets is minimal.}
    \label{tab:same_train_data_comparison}
\end{table*}

\subsection{Settings}\label{sec:settings}

\paragraph{Training details.}
SLIMER is based on LLaMA-2 7B chat~\cite{touvron2023llama}. Investigating how different families or model sizes affect results is outside the objectives of our work.  
In all the experiments the models were fine-tuned with LoRA~\citep{hu2021lora} $r=8$, $\alpha=16$ for 10 epochs with early stopping, 32 batch size and learning rate initialized to $3.0 \times 10^{-4}$ with cosine scheduler and a warm-up of 60 steps. Context length was set to 768, longer inputs were chunked.
In order to reduce the overlap between training and test entities (see Table~\ref{tab:NEs_overlap}), we trained SLIMER on the PileNER-type subset described in Section~\ref{subsec:exps_datasets}, which comprises 391 distinct entity types. As already observed in literature, instruction-tuning helps aligning the model to the tasks, and most of the gains in performance are achievable with little amounts of instructions~\citep{wang2022self, zhou2024lima, zugarini-etal-2024-clue-instruct}. Hence, we only picked 5 examples per class from the PileNER-type subset. In addition, for each example containing at least one annotated instance, we also included a negative example, i.e. one without any named entity. Overall, SLIMER was trained on 3910 input sequences, a small fraction of the data typically fed to other state-of-the-art models. To better understand the training set scales of different models, in Figure~\ref{fig:x_vs_y_training_samples} we depicted them with circles proportional to their training sizes.


\paragraph{Metrics.} We align with existing work in the literature by computing the F1 metric in a strict manner, i.e. for a given NE type, all the unique text spans within the text passage are required to be retrieved, each with all its associated tokens and no additional ones added. 
Any reported SLIMER result is the averaged value of three different runs of the same training-evaluation configuration. Hence, we also provide the standard deviation of our experiments.

\subsection{Compared Models}
\label{sec:compared_models}
We take into account the several state-of-the-art approaches for zero-shot NER:\\

\textbf{ChatGPT}~\citep{brown2020language}, prompted with the same strategy in \citet{ye2023comprehensive}. While it constitutes a baseline not specifically instructed for NER, it serves as a reference baseline to demonstrate the out-of-the-box performance of a pre-trained LLM, appropriately prompted for NER, as done in previous works \cite{zhou2023universalner, zaratiana2023gliner, ding2024rethinking}. 

\textbf{InstructUIE}~\citep{wang2023instructuie}, based on the flan-t5-xxl encoder-decoder architecture of 11B parameters.

\textbf{UniNER}~\citep{zhou2023universalner}, a family of LLMs all based on the LLaMA-1-7B. Despite being an older backbone, authors themselves observed no significant improvements when experimenting their approach with LLaMA-2. Therefore, we keep their original model.
We evaluate UniNER's three variants: \textit{type} is trained on full PileNER-type, described in Subsection~\ref{subsec:exps_datasets}; \textit{type+sup.} was trained also on a collection of human-labeled NER datasets; \textit{def}, on the other hand, differs from \textit{type} by the presence of short definitions of the entity type, as described in Section \ref{sec:related_works}.

\textbf{GoLLIE} ~\citep{sainz2024gollie} is a model based on Code-LLaMA leveraging annotation guidelines formatted in a code-like representation. In the comparison we considered only the 7B version, since it has a number of parameters similar to most of the other approaches, SLIMER included.

\textbf{GLiNER-L}~\citep{zaratiana2023gliner} is an encoder-only DeBERTa-304M parameters model. We choose their biggest and performing model. Nonetheless, GLiNER-L is by far the smallest one amongst the selected state-of-the-art approaches, yet its quite competitive on OOD zero-shot NER.

\textbf{GNER}~\citep{ding2024rethinking} was released in two versions differing for their backbone LLM: flan-t5-xxl and LLaMA-7B. They are referred as GNER-T5 and GNER-LLaMA, respectively.

\subsection{Off-the-Shelf Models for Zero-shot NER} 
We compare SLIMER against state-of-the-art models on the benchmarks described in Section~\ref{subsec:exps_datasets}.

\paragraph{Out-of-Domain (OOD) inputs.} Table~\ref{tab:MIT_CrossNER_comparison} compares SLIMER against state-of-the-art models on MIT and CrossNER benchmarks. We consider this benchmark as OOD, since most of the existing models have been trained on examples including most of the entity classes present in the test set (see last column and Table~\ref{tab:NEs_overlap}). However, our model has minimal training/test classes overlap, thus operating in a more zero-shot scenario. Despite that, and the fact that we used only a fraction of the training data with respect to other models, SLIMER offers competitive performance, surpassing several of the existing state-of-the-art models. Moreover, our training data is entirely synthetic, whereas models like GoLLIE or UniNER-type+sup also exploit human-annotated text. The importance of gold-annotated examples can be depicted by observing the 16\% absolute increase between UniNER-type and UniNER-type+sup (see Table~\ref{tab:MIT_CrossNER_comparison}).


\begin{table}[htb]
\centering
\small
\resizebox{\columnwidth}{!}{
\begin{tabular}{l|cc}
    \toprule
    \textbf{Model} & $\mu$-\textbf{F1} & M-\textbf{F1}\\ 
    
    \midrule
    
    UniNER-type & 34.78 & \underline{37.58}\\ 
    
    UniNER-def & 33.62 & 31.80\\ 

    UniNER-type+sup. & \underline{37.82} & 36.79 \\ 

    GoLLIE \dag & 27.68 & 24.13\\

    GLiNER-L & 26.57 & 24.34\\ 

    GNER-T5 & 27.88 & 30.26\\ 

    GNER-LLaMA & 23.58 & 14.21\\ 

    \midrule
    SLIMER w/o D\&G & $40.41\pm{5.09}$ & $35.90\pm{4.08}$\\
    
    \textbf{SLIMER} \dag & $\textbf{45.27}\pm\textbf{1.04}$ & $\textbf{40.14}\pm\textbf{0.44}$\\ 
        
    \bottomrule
    \end{tabular}
}
\caption{Comparing SLIMER against SOTA models on BUSTER to assess generalization over never-seen-before NEs. Models leveraging on guidelines are denoted with symbol \dag.
} \label{tab:BUSTER_comparison}
\end{table}

\begin{figure}[htb]
\centering
\includegraphics[width=1\columnwidth]{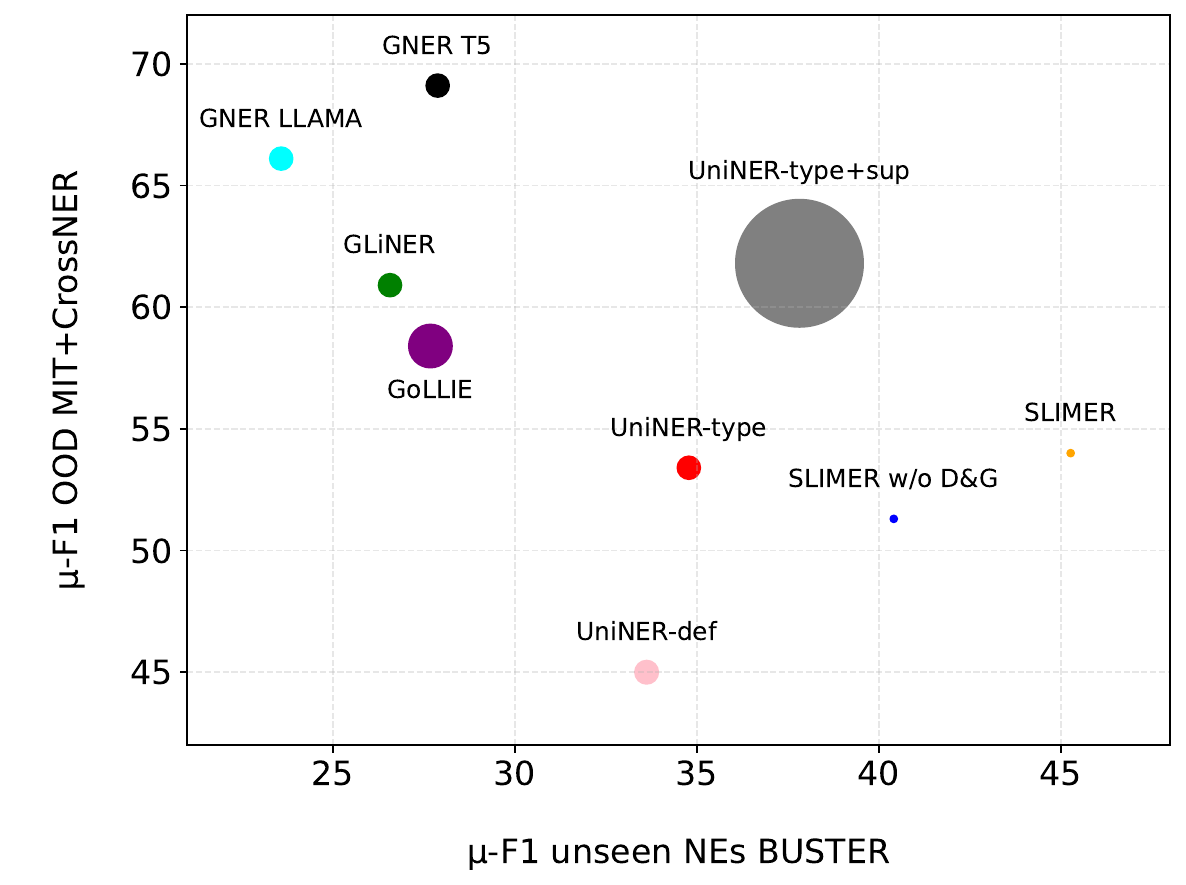}
  \caption{Comparing SOTA models: $\mu$-F1 scores on unseen NEs in BUSTER (x-axis), OOD evaluation on MIT/CrossNER (y-axis). Circles' size is proportional to the number of examples seen in training by each model.}
  \label{fig:x_vs_y_training_samples}
\end{figure}

\paragraph{Never-Seen-Before NEs.} To experiment the ability of existing models on never-seen-before labels, we extend the evaluation on BUSTER, which is characterized by financial entities that are rather far from the more traditional tags observed by all models during training. For simplicity, we limited the evaluation to most performing approaches only, thus we omitted ChatGPT and InstructUIE in this experimentation.
Results outlined in Table~\ref{tab:BUSTER_comparison} exhibit an inverse trend with respect to OOD experiments. Indeed, best scoring state-of-the-art models in MIT and and CrossNER, such as the ones from GNER family, both under-perform in BUSTER, also due to their inability to work on long input texts\footnote{Except for GoLLIE, all the other approaches require input chunking into multiple smaller passages. While the sliding window for most is set to 900 words, for GNER we are limited to work on 150 words per input text.}.
Analogously, GLiNER and GoLLIE struggle in such a benchmark. Notably, GoLLIE's performance is unexpectedly low, given that it is the only other to exploit annotation guidelines and it was fed with the same guidelines we provide to our SLIMER. We speculate the CodeLLaMA backbone is less familiar with financial domain inputs and such long definition and guidelines. SLIMER instead, appears to be the most effective in dealing with unseen labels, thanks to its lighter instruction-tuning methodology. Figure~\ref{fig:x_vs_y_training_samples} clearly delineates such a behaviour.

{\andrew
\label{par:same_framework_comparison}
\subsection{Comparing Zero-shot NER Approaches} 
Existing approaches are all trained under different conditions (e.g. backbone, training data, entities overlap), thus making difficult to evaluate the approach per se. 
To isolate the methodology from the rest, we compare them under the same conditions. We use the same backbone (LLaMA2-chat-7B) and fine-tune all the models as described in Section~\ref{sec:settings}. Such a setting also allows a better evaluation of actual zero-shot performance of existing models. 
In this experiment, we selected the best two performing approaches, namely GNER and GoLLIE\footnote{Further details are provided in Appendix~\ref{app:investigative_experiments}.}. 

Results are reported in Table~\ref{tab:same_train_data_comparison}. We found that, without an extensive training corpus covering most of the entity classes, GNER performance drops of 18 F1 points, scoring now 6 points below SLIMER. Additionally, its fine-tuning required an increase in the number of samples per entity type from 5 to 50, further revealing its high dependency on training data. GoLLIE, on the other hand, better preserves its generalization capabilities to unseen entity types, scoring 1 point less than SLIMER in MIT and CrossNER. Moreover, the gap between SLIMER and GoLLIE in BUSTER further increases. 
Notably, GoLLIE's performance decrease is not solely due to the new training corpus, but also to the use of a non-code-oriented backbone\footnote{Using its original CodeLLaMA model would improve performance by 3\%, as shown in Table~\ref{tab:GOLLIE_degradation}, Appendix \ref{app:investigative_experiments}.}. This limits the applicability of such an approach to non-code-oriented LLMs, as noted by the authors themselves.

}

\begin{table*}[ht!]
  \centering
  \tiny
  \resizebox{\textwidth}{!}{
  \begin{tabular}{lc|p{12cm}|cc|c}

    \toprule
    \textbf{Dataset} & \textbf{NE} & \multicolumn{1}{c|}{\textbf{Definition \& Guidelines}} & \textbf{SLIMER w/o D\&G F1} & \textbf{SLIMER-F1} & \textbf{$\Delta$} \textbf{F1}\\
    \midrule





    \rowcolor{LightGreen}
    Movie & Trailer & Definition: 'trailer' refers to a short promotional video that provides a preview or teaser of a forthcoming movie., Guidelines: Label also general entertainment terms like 'preview' or 'teaser'., & 23.44 & 58.62 & +35.18 \\

    \midrule
    
    \rowcolor{LightCyan}
    Science & Chemical Compound & Definition: 'chemical compound' refers to distinct chemical substances composed of two or more elements in fixed proportions., Guidelines: Label entities as 'chemical compound' if they are not proteins or enzymes. Exercise caution with ambiguous terms like 'Almond', which can refer to both a food item and a chemical compound (benzaldehyde). Be aware of complex nomenclature and chemical structures when identifying compounds. & 50.32 & 58.85 & +8.53 \\

    \midrule

    \rowcolor{LightRed}
    Restaurant & Amenity & Definition: 'amenity' refers to services, facilities, or features that enhance the convenience, comfort, or enjoyment of a location., Guidelines: When annotating 'amenity', focus on tangible or accessible services and facilities. Avoid labeling abstract concepts, such as 'ambiance' or 'vibe', that are not clearly associated with a specific amenity. Examples of 'amenity' are 'steampunk flavored', 'upscale place' and 'reservation'. & 33.38 & 28.18 & -5.20 \\

\bottomrule
  \end{tabular}
}
  \caption{
    Some examples of synthetically generated definition and guidelines. Absolute F1 gains between SLIMER and its version without definition and guidelines are reported. 
  }\label{tab:SLIMER_examples}
\end{table*}

\subsection{Ablation studies} We conducted ablation studies to understand how zero-shot NER performance is affected as training data size increases, with or without the inclusion of definition and guidelines.

\begin{figure}[htb]
  \centering
  \includegraphics[width=1\columnwidth]{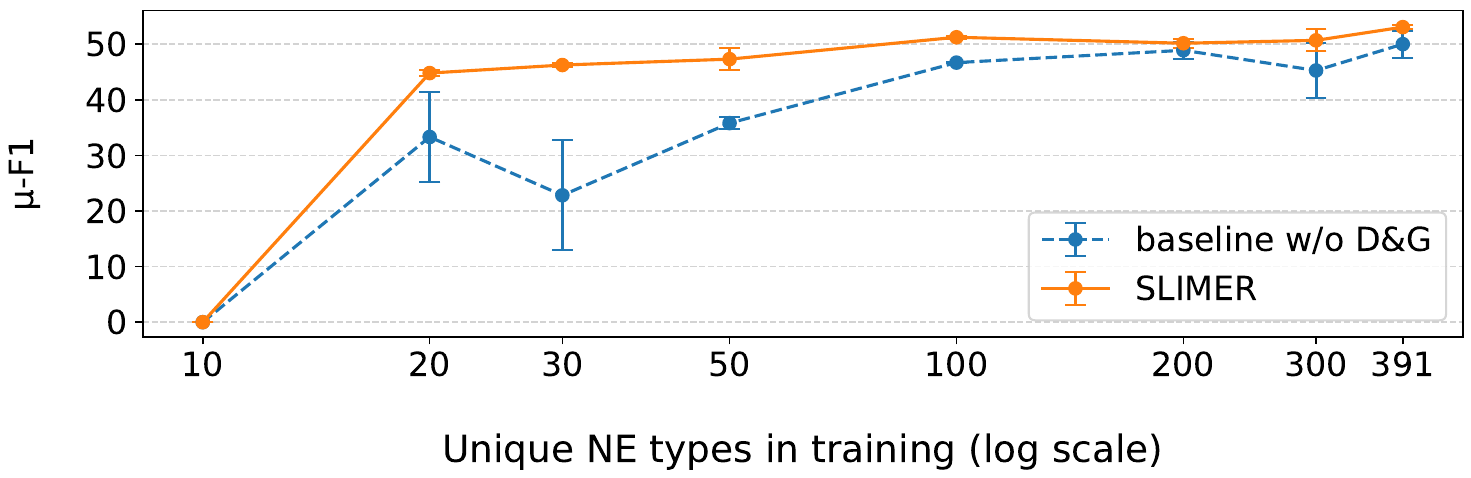} \\[1ex]
  \includegraphics[width=1\columnwidth]{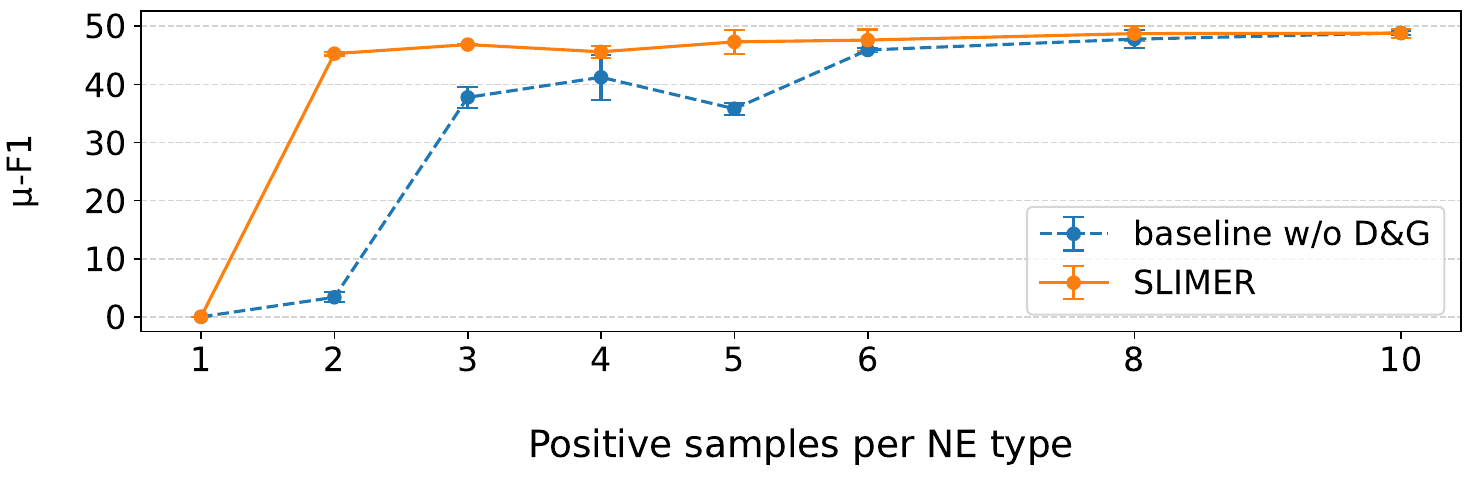}
  \caption {$\mu$ F1 scores of SLIMER and its baseline without D\&G on MIT, CrossNER and BUSTER altogether, as we increase the number of unique NEs (top) and the number of samples per NE (bottom) seen in training.}
  \label{fig:growing_training_data}
\end{figure}

\paragraph{Increasing training data.} 
We investigated how zero-shot NER performance changes as the number of training instances increases. We proceeded in two directions: (1) increasing the number of unique NEs tags, while sampling a fixed total amount of samples per type; (2) varying the number of examples per NE, while keeping fixed the number of distinct categories. In particular, we set 10 examples (5 positive and 5 negative) per class in (1), and we set the number of distinct NEs to 50 in experiment (2). In both cases, we progressively add new classes/examples to the already selected ones.
F1 scores are measured by considering MIT, CrossNER and BUSTER as a unique benchmark. By observing the results outlined in Figure~\ref{fig:growing_training_data}, it emerges that increasing the number of unique entity types somewhat improves results, albeit most of the progresses are achievable with just 20 entity tags.
Notably, the poor scores achieved by the model when trained on very few examples, are not due to the model's inability to identify entities, but rather to its struggles in aligning the output to the required format.
Adding more than two examples per class brings little to no benefits, instead. This confirms what already observed in literature~\citep{wang2022self,zhou2024lima}, further motivating our decision to instruct models on a small portion of the training data. 

\paragraph{Impact of Definition and Guidelines.} As a final analysis, we assess the impact of definition and guidelines. We compare SLIMER with a version of it devoid of definition and guidelines, referred as SLIMER w/o D\&G. Results, reported in tables~\ref{tab:MIT_CrossNER_comparison},\ref{tab:BUSTER_comparison} and Figure~\ref{fig:growing_training_data} consistently demonstrate how the definition and guidelines are helpful to the model. Indeed, there are improvements in F1 in both OOD and never-seen-before scenarios. Moreover, the absence of guidelines also significantly increases the standard deviation over multiple runs, thus demonstrating that D\&G also make the learning more consistent and stable. Notably, we can also observe from Figure~\ref{fig:growing_training_data} that in order to reach comparable performance, SLIMER w/o D\&G requires more training data. To better understand how definition and guidelines contribute in improving the model, we show some qualitative examples in Table \ref{tab:SLIMER_examples}\footnote{Additional examples can be seen in Appendix~\ref{app:appendix_some_examples}.}. For each example, we report F1 scores obtained by using SLIMER or its version lacking of D\&G. 
Information about an entity type not only helps in detecting novel NEs, but it can also be beneficial to disambiguate polysemous tags, such as ``trailer''. 

\section{Conclusions}
In this paper, we presented SLIMER, an instruction-tuned LLM for zero-shot NER. 
With a prompt enriched with definition and annotation guidelines, and a fine-tuning on a restricted set of entity tags, SLIMER, differently from most of the existing models, is specifically designed to better deal with unseen named entity tags. Experiments show that definition and guidelines steer the annotation process, especially on never-seen-before classes, thus yielding better predictions and a more stable learning. Furthermore, SLIMER performs comparably to state-of-the-art approaches, while being trained on a fraction of samples and entity types having little overlap with the test set. 
In the future, we plan to broaden the scope of SLIMER to any Information Extraction problem. Moreover, we will investigate solutions to better scale to large sets of entity classes to tag.

\clearpage
\newpage

\section*{Limitations}
A primary limitation of our approach lies in the instruction-tuning template we have adopted. While extracting the occurrences of a named entity per prompt allows for shorter instructions and a stronger focus on the definitions and guidelines components, it results in the overhead of requiring a number of inference calls per input text equal to the cardinality of the label set. Consequently, our approach does not scale well on datasets with a large number of entity classes.

Another potential limitation could stem from data contamination between the benchmark datasets and the pre-training data of the LLM. However, any performance gap between SLIMER and its baseline can be attributed to the presence of the D\&G additional components, as both models share the same pre-training data.

Finally, in a real-world scenario, the three examples required to automatically generate definitions and guidelines might not be available for new unseen labels. Nevertheless, in practice, manually preparing three examples per entity class may be still easier than creating detailed definitions and guidelines for a human annotator.

\section*{Acknowledgments}
\label{sec:funding}

The work was partially funded by:
\begin{itemize}
    \item ``ReSpiRA - REplicabilità, SPIegabilità e Ragionamento'', a project financed by FAIR, Affiliated to spoke no. 2, falling within the PNRR MUR programme, Mission 4, Component 2, Investment 1.3, D.D. No. 341 of 03/15/2022, Project PE0000013, CUP B43D22000900004  \footnote{RESPIRA: \url{https://www.opencup.gov.it/portale/web/opencup/home/progetto/-/cup/B43D22000900004}};
    \item ``MAESTRO - Mitigare le Allucinazioni dei Large Language Models: ESTRazione di informazioni Ottimizzate'' a project funded by Provincia Autonoma di Trento with the Lp 6/99 Art. 5:ricerca e sviluppo, PAT/RFS067-05/06/2024-0428372, CUP: C79J23001170001  \footnote{MAESTRO: \url{https://www.opencup.gov.it/portale/web/opencup/home/progetto/-/cup/C79J23001170001}};
    %
    %
    \item ``enRichMyData - Enabling Data Enrichment Pipelines for AI-driven Business Products and Services'', an Horizon Europe (HE) project, grant agreement ID: 101070284 \footnote{\url{https://doi.org/10.3030/101070284}}.
\end{itemize}
%


\bibliography{custom}

\clearpage

\appendix

\section{Qualitative Analysis}
\label{app:appendix_some_examples}
In Table~\ref{tab:perNE_scores} we report the F1 scores for some NE classes with the purpose of getting a better insight into the usefulness of the definitions and guidelines for zero-shot NER. We aim to list potential benefits these components can provide and support our thesis with some examples. However, the examples provided are not intended for quantitative assessment, rather they serve as illustrative instances supporting some of our claims.

\begin{table*}[htb]
  \centering
  \tiny
  \resizebox{\textwidth}{!}{
  \begin{tabular}{lc|p{12cm}|cc|c}

    \toprule
    \textbf{Dataset} & \textbf{NE} & \multicolumn{1}{c|}{\textbf{Definition \& Guidelines}} & \textbf{baseline-F1} & \textbf{SLIMER-F1} & \textbf{$\Delta$} \textbf{F1}\\
    \midrule

    \rowcolor{LightGreen}
    BUSTER & Generic Consulting Company & Definition: 'generic consulting company' refers to a business entity that provides non-legal advisory services in areas such as finance, accounting, due diligence, and other professional consulting services., Guidelines: Avoid labeling a company that primarily provides legal services. Exercise caution with company names that include personal names which might be confused with individuals, and consider the context to determine whether the reference is to a company. & 2.34 & 14.78 & +12.44 \\

    \midrule
    
    \rowcolor{LightCyan}
    BUSTER & Selling Company & Definition: 'selling company' refers to a company that is selling or divesting assets, subsidiaries, or equity to another party as part of a transaction., Guidelines: Be careful when identifying the entity actually doing the selling, as it may not be the main subject of the sentence or document. Pay attention to possessive forms and synonyms such as 'vendor', 'owner', or 'parent company'. The company's role as 'seller' must be understandable from the sentence in which it appears. Do not label company names which role in the transaction is not evident. & 21.13 & 31.04 & +9.91 \\
    
    \midrule
    
    \rowcolor{LightCyan}
    BUSTER & Buying Company & Definition: 'buying company' refers to a company that is acquiring another company or its assets through a transaction or merger., Guidelines: When recognizing 'buying company' entities, focus on the company names directly involved in the acquisition process as buyers, while being careful not to label subsidiaries or companies in other roles. The company's role as 'buyer' must be understandable from the sentence in which it appears. Do not label company names which role in the transaction is not evident. & 43.09 & 49.84 & +6.75 \\

    \midrule

    \rowcolor{LightCyan}
    AI & Person & Definition: 'person' refers to individuals, including public figures, celebrities, and notable personalities., Guidelines: If a person is working on research (including professor, Ph.D. student, researcher in companies, and etc) avoid labeling it as 'person' entity. & 35.50 & 41.15 & +5.65 \\

    \midrule

    \rowcolor{LightGreen}
    AI & University & Definition: 'university' represents educational institutions that offer higher education and academic research programs., Guidelines: Avoid labeling general concepts such as 'education' or 'academia' as 'university'. Exercise caution with ambiguous terms like 'Cambridge' (can refer to different institutions) and 'Harvard' (can refer to a person). & 64.24 & 78.38 & +14.14 \\
    
    \midrule

    \rowcolor{LightRed}
    AI & Product & Definition: 'product' refers to tangible or intangible items, systems, or tools, including but not limited to physical products, software, and industrial machinery, designed for specific functions or applications., Guidelines: Exercise caution when dealing with ambiguous terms like 'Java' (programming language, island, and coffee). Consider the context to discern the correct entity. Be mindful of generic terms like 'system' and 'toolkit' which may require additional context to determine if they fall under the 'product' category. & 20.36 & 6.02 & -14.33 \\

    \midrule

    \rowcolor{LightGreen}
    Literature & Event & Definition: 'event' refers to specific incidents, occurrences, or happenings that take place at a particular time and location, such as festivals, wars, conferences, and award ceremonies., Guidelines: Avoid labeling general or ongoing occurrences, such as 'daily routine' or 'regular meetings'. Exercise caution with ambiguous terms like 'revolution' (can refer to a political event or a spinning motion) and 'strike' (can denote a labor event or a military action). & 37.89 & 54.35 & +16.47 \\

    \midrule

    \rowcolor{LightCyan}
    Movie & Genre & Definition: 'genre' refers to a category or classification characterized by specific stylistic, thematic, or content elements., Guidelines: Avoid labeling general terms like 'film', 'book', or 'music' as 'genre'. Focus on labeling genres such as 'science fiction', 'romance', 'horror' and similar. Exercise caution with ambiguous terms that can belong to multiple genres, such as 'drama' (which can refer to a genre or a situation within a story)., & 38.39 & 46.97 & +8.58 \\

    \midrule

    \rowcolor{LightGreen}
    Movie & Trailer & Definition: 'trailer' refers to a short promotional video that provides a preview or teaser of a forthcoming movie., Guidelines: Label also general entertainment terms like 'preview' or 'teaser'., & 23.44 & 58.62 & +35.18 \\

    \midrule
    
    \rowcolor{LightCyan}
    Movie & Title & Definition: 'title' refers to names of creative works, such as movies, books, music albums, and artistic productions., Guidelines: Avoid labeling generic words that can be interpreted as common nouns, like 'run' or 'the'. Exercise caution with potentially ambiguous cases, such as 'Savannah Hilton' (person with a title name) or 'Apple' (brand and fruit). & 31.42 & 33.01 & +1.58 \\

    \midrule

    \rowcolor{LightGreen}
    Politics & Person & Definition: 'person' refers to individual human beings, but who are not politicians., Guidelines: In this political context, avoid labelling as 'person' people who are 'politicians'., & 53.87 & 71.47 & +17.60 \\

    \midrule

    \rowcolor{LightRed}
    Music & Award & Definition: 'award' refers to a recognition or honor given to individuals, groups, or organizations in various fields, such as sports, entertainment, academia, and business., Guidelines: Avoid labeling non-official titles or generic terms like 'best', 'top', 'favorite'. Exercise caution with ambiguous terms like 'Oscar' (could refer to a person or the award) or 'Nobel' (could refer to the organization or the prize). & 67.58 & 62.27 & -5.31 \\

    \midrule

    \rowcolor{LightCyan}
    Music & Song & Definition: 'song' refers to a musical composition with lyrics and melody that can be performed or recorded., Guidelines: Do not label general music-related terms like 'album' or 'lyrics'. Exercise caution with ambiguous cases like 'Billie Jean' (which can refer to a person or a song) or 'Guns N' Roses' (which can refer to a band or a song)., & 59.56 & 67.39 & +7.83 \\

    \midrule
    
    \rowcolor{LightGreen}
    Restaurant & Price & Definition: 'price' represents the cost or value of a product or service in a given context., Guidelines: Also consider labeling terms like 'cheap', 'inexpensive' and similar, when referring to a product or service., & 30.41 & 42.96 & +12.55 \\

    \midrule
    
    \rowcolor{LightRed}
    Restaurant & Amenity & Definition: 'amenity' refers to services, facilities, or features that enhance the convenience, comfort, or enjoyment of a location., Guidelines: When annotating 'amenity', focus on tangible or accessible services and facilities. Avoid labeling abstract concepts, such as 'ambiance' or 'vibe', that are not clearly associated with a specific amenity. Examples of 'amenity' are 'steampunk flavored', 'upscale place' and 'reservation'. & 33.38 & 28.18 & -5.20 \\

    \midrule
    
    \rowcolor{LightGreen}
    Science & Astronomical Object & Definition: 'astronomical object' refers to celestial bodies, such as planets, moons, asteroids, and comets, that exist in outer space., Guidelines: Avoid labeling general terms like 'orbit', 'gravitation', or 'gravity assist'. Exercise caution with terms that can have multiple meanings, such as 'Moon' (natural satellite vs. a generic noun) or 'Mars' (the planet vs. the god of war). & 40.18 & 50.75 & +10.57 \\

    \midrule
    
    \rowcolor{LightCyan}
    Science & Chemical Compound & Definition: 'chemical compound' refers to distinct chemical substances composed of two or more elements in fixed proportions., Guidelines: Label entities as 'chemical compound' if they are not proteins or enzymes. Exercise caution with ambiguous terms like 'Almond', which can refer to both a food item and a chemical compound (benzaldehyde). Be aware of complex nomenclature and chemical structures when identifying compounds. & 50.32 & 58.85 & +8.53 \\

    \bottomrule
  \end{tabular}
}
  \caption{
    Comparing SLIMER to its baseline w/o D\&G on some Named Entities. In red are performance degradations, in blue are positive improvements between 0 and 10, in green are very high improvements over 10 points.
  }\label{tab:perNE_scores}
\end{table*}

\paragraph{Different granularity and exceptions.} ``Every musician is also a person''. However, as occurs in CrossNER, there are cases where an individual should be labelled as a person only if it does not fall into the categories of musician, scientist, writer or politician. From Table~\ref{tab:perNE_scores} we can effectively see how the guidelines instruct SLIMER with such requirements and the model improves in both precision and final F1 with respect to its baseline.

\paragraph{Different annotation schemes.} Guidelines are key for flexibility to new annotation schemes. As often the case in Zero-Shot NER, a NE in test may require to include or exclude particular instances with respect to what has been trained on; similarly, in supervised-setting, different datasets may adopt different annotation schemes for the same NE (e.g., \citet{zhou2023universalner} are required to specify in the prompt to which dataset the sample belongs). By carefully formulating the NE definition and guidelines, we can flexibly adapt to the desired behaviour. However, as we can see from the red cases, this can sometimes lead to a drop in performance. We believe that this can be partly due to overly strict guidelines, resulting in higher precision at the expense of lower recall.

\paragraph{Polysemous Named Entities.} Guidelines potentially solve the problem of polysemous NE types where, for example, the same NE ``title'' may denote film titles or nobility titles. We briefly experiment by leaving the NE ``title'' in the PileNER training set (where it denotes nobility titles) and evaluating on the NE title from MIT-Movie dataset. However, the improvement is only of +2 points, probably because the backbone model is already somehow able to adapt to the correct sense given the context. On the opposite, the polysemous tag ``trailer'' benefits from the D\&G.

\paragraph{Provide external knowledge.} Finally, and most importantly in Zero-Shot NER, annotation guidelines may enable the labelling of never-before-seen Named Entities based on the model's ability to adhere to the provided guidelines, thus acting as a source of external knowledge for the model.


\begin{figure}[htb]
\centering
  \includegraphics[width=0.8\columnwidth]{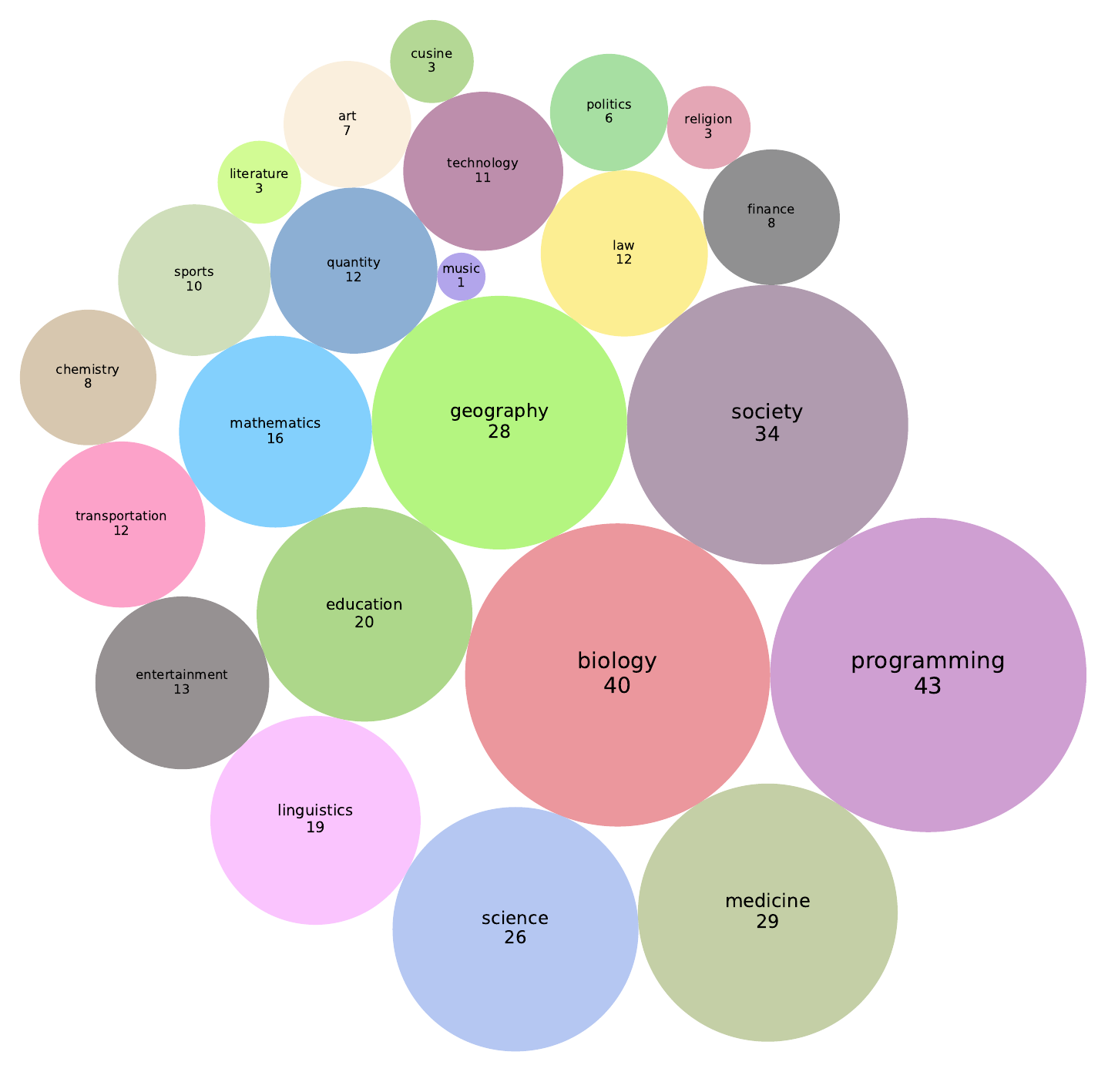}
  \caption{The 391 Named Entities in the PileNER-type subset, grouped by macro topics. ``misc'' (not shown) groups 26 NEs that do not fit into the defined topics.}
  \label{fig:macro-topics}
\end{figure}

\newpage
\section{Further experiments}
\label{app:investigative_experiments}

\begin{table*}[htb]
\centering
\footnotesize
\resizebox{\textwidth}{!}{
\begin{tabular}{llc|cc|ccccc|c||c}
    \toprule
    
    
    \textbf{Model} & \textbf{Backbone} & \textbf{Guidelines} &
    \multicolumn{2}{c|}{\textbf{MIT}} & \multicolumn{5}{c|}{\textbf{CrossNER}} & & \textbf{BUSTER}\\

    \textbf{} & \textbf{} & & Movie & Restaurant & AI & Literature & Music & Politics & Science & AVG & \textbf{} \\

    \midrule
    
    GoLLIE & CodeLLaMA-7B & $\checkmark$ & 53.9 & 45.4 & 54.9 & 56.8 & 66.5 & 64.2 & 58.0 & 57.1 & 14.5\\

    GoLLIE & LLaMA2-chat-7B & $\checkmark$ & 47.9 (-6.0) & 39.2 (-6.2) & 54.2 (-0.7) & 51.4 (-5.4) & 62.5 (-4.0) & 61.9 (-2.3) & 55.0 (-3.0) & 53.2 (-3.9) & 15.2 (+0.7)\\

    \bottomrule
    \end{tabular}
}
    \caption{Instruction-tuning GoLLIE on the same sub-set of unique Named Entity types of SLIMER, using its original code-oriented LLM and a non-code-oriented LLM as the one employed by SLIMER.}
    \label{tab:GOLLIE_degradation}
\end{table*}

\paragraph{GNER and GoLLIE training details.} To ensure reproducibility, we here detail the training setups used for GNER and GoLLIE, implemented using the same LLM backbone and subset of unique entity types of SLIMER. While GoLLIE fine-tuned effectively on same SLIMER's training data, GNER required significantly more samples per entity (50 instead of 5) to achieve comparable performance. While adding more examples (100 per NE) did not improve performance, this highlighted GNER's dependence on training data.

Re-implementing both GNER and GoLLIE approaches was easily done using their provided scripts and code hosted in their respective Github repositories. We kept the same training hyper-parameters in their original scripts, as described in their respective papers \cite{ding2024rethinking, sainz2024gollie}. We only increased the number of training epochs to 10, to align all the models with SLIMER. 

Regarding GNER, because it's not feasible to list all 391 NEs of PileNER-type subset in the prompt, for each input text we randomly sampled 10 entity types, ensuring that the positive tags in the passage of text were included. Regarding GoLLIE, we converted PileNER-subset to GoLLIE's format by similarly sampling 10 total guidelines per input, ensuring the positive where included. We reduced the label noise to 0.2 and guidelines dropout to 0.1, because of the reduced number of training samples, while we did not rephrase the guidelines. Indeed, we provided the same definition and guidelines of SLIMER, formatting them as doc-strings.

\paragraph{GoLLIE-CodeLLaMA on PileNER-subset.} We conducted further experiments to the ones in Section~\ref{par:same_framework_comparison} by implementing GoLLIE on the same training data as SLIMER, using its original code-oriented backbone. Results in Table~\ref{tab:GOLLIE_degradation} show a 4-point drop in performance when using a non-code-oriented backbone. This underlines a significant limitation in GoLLIE's adaptability, as noted by \citet{sainz2024gollie} themselves, suggesting reduced applicability of the approach to LLMs that are not code-oriented.



\end{document}